\documentclass[10pt,twocolumn,letterpaper]{article}

\usepackage{cvpr}              

\usepackage{graphicx}
\usepackage{amsmath}
\usepackage{amssymb}
\usepackage{booktabs}
\usepackage[table]{xcolor}
\usepackage{pifont}
\usepackage{bm}
\usepackage{multirow}
\usepackage{rotating}
\usepackage{siunitx}

\usepackage[colorlinks,pagebackref,urlcolor=purple, citecolor=blue,bookmarks=false,hypertexnames=true]{hyperref}

\usepackage[capitalize]{cleveref}
\crefname{section}{Sec.}{Secs.}
\Crefname{section}{Section}{Sections}
\Crefname{table}{Table}{Tables}
\crefname{table}{Tab.}{Tabs.}

\def\@onedot{\ifx\@let@token.\else.\null\fi\xspace}
\def\eg{\emph{e.g}\onedot} 
\def\ie{\emph{i.e}\onedot}

\def\cvprPaperID{4939} 
\def\confName{CVPR}
\def\confYear{2023}

\definecolor{darkgreen}{RGB}{0, 150, 0}
\definecolor{darkred}{RGB}{200, 0, 0}
\definecolor{darkblue}{RGB}{0, 0, 200}
\newcommand{\ch}{{\color{darkgreen} \ding{51}}}
\newcommand{\xm}{{\color{darkred} \ding{55}}}

\definecolor{gray9}{gray}{.9}
\definecolor{gray95}{gray}{.95}
\definecolor{gray8}{gray}{.8}
\definecolor{gray85}{gray}{.85}
\newcommand{\nap}{{\color{gray} NA}}
\newcommand{\CC}[1]{\cellcolor{black!#1}\:\!\!\!}

\begin{document}

\title{\textbf{X$^3$}KD: Knowledge Distillation Across Modalities, Tasks and Stages \\ for Multi-Camera 3D Object Detection}

\author{
Marvin Klingner \thanks{These authors contributed equally to this work.} $^{,}$\thanks{Automated Driving, Qualcomm Technologies, Inc.}
\and 
Shubhankar Borse \footnotemark[1] $^{,}$\thanks{Qualcomm AI Research, an initiative of Qualcomm Technologies, Inc.}
\and
Varun Ravi Kumar \footnotemark[1] $^{,}$\footnotemark[2]
\and
Behnaz Rezaei \footnotemark[2]
\quad\quad
Venkatraman Narayanan \footnotemark[2]
\quad\quad
Senthil Yogamani \thanks{Automated Driving, QT Technologies Ireland Limited}
\quad\quad
Fatih Porikli \footnotemark[3] \\
{\tt\small \{mklingne, sborse, vravikum, brezaei, vennara, syogaman, fporikli\}@qti.qualcomm.com}\\
}
\maketitle
\begin{abstract}\vspace{-6pt}
Recent advances in 3D object detection (3DOD) have obtained remarkably strong results for LiDAR-based models. In contrast, surround-view 3DOD models based on multiple camera images underperform due to the necessary view transformation of features from perspective view (PV) to a 3D world representation which is ambiguous due to missing depth information. This paper introduces X$^3$KD, a comprehensive knowledge distillation framework across different modalities, tasks, and stages for multi-camera 3DOD. Specifically, we propose cross-task distillation from an instance segmentation teacher (X-IS) in the PV feature extraction stage providing supervision without ambiguous error backpropagation through the view transformation. After the transformation, we apply cross-modal feature distillation (X-FD) and adversarial training (X-AT) to improve the 3D world representation of multi-camera features through the information contained in a LiDAR-based 3DOD teacher. Finally, we also employ this teacher for cross-modal output distillation (X-OD), providing dense supervision at the prediction stage. We perform extensive ablations of knowledge distillation at different stages of multi-camera 3DOD. Our final X$^3$KD model outperforms previous state-of-the-art approaches on the nuScenes and Waymo datasets and generalizes to RADAR-based 3DOD. Qualitative results video at {\href{https://youtu.be/1do9DPFmr38}{https://youtu.be/1do9DPFmr38}}.
\end{abstract}
\vspace{-18pt}
\section{Introduction}
\label{sec:intro}
\vspace{-3pt}

\begin{figure}
	\captionsetup{font=small, belowskip=-16pt}
    \centering
	\includegraphics[width=0.95\linewidth]{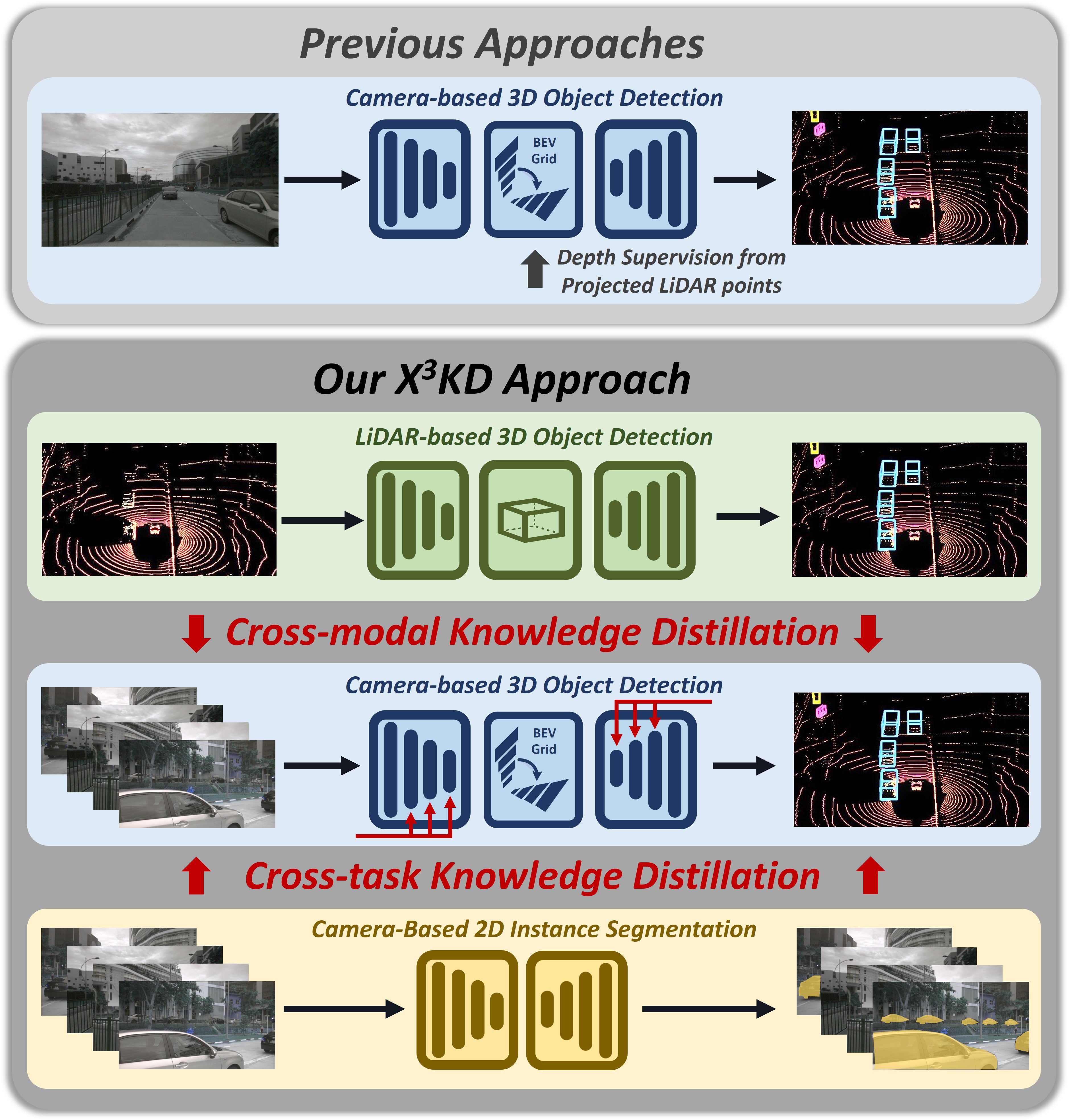}
    \vspace{-5pt}
    \caption{While previous approaches considered multi-camera 3DOD in a standalone fashion or with depth supervision, \textbf{we propose X$^3$KD, a knowledge distillation framework using cross-modal and cross-task information} by distilling information from LiDAR-based 3DOD and instance segmentation teachers into different stages (marked by red arrows) of the multi-camera 3DOD.}
    \label{fig:concept}
\end{figure}
3D object detection (3DOD) is an essential task in various real-world computer vision applications, especially autonomous driving. Current 3DOD approaches can be categorized by their utilized input modalities, \eg, camera images~\cite{li2022bevdepth, philion2020lift, wang2022detr3d} or LiDAR point clouds~\cite{lang2019pointpillars, yin2021center, zhou2018voxelnet}, which dictates the necessary sensor suite during inference. Recently, there has been significant interest in surround-view multi-camera 3DOD, aiming to leverage multiple low-cost monocular cameras, which are conveniently embedded in current vehicle designs in contrast to expensive LiDAR scanners. Existing solutions to 3DOD are mainly based on extracting a unified representation from multiple cameras \cite{li2022bevformer, li2022bevdepth, park2021pseudo, reading2021categorical} such as the bird's-eye view (BEV) grid. However, predicting 3D bounding boxes from 2D perspective-view (PV) images involves an ambiguous 2D to 3D transformation without depth information, which leads to lower performance compared to LiDAR-based 3DOD~\cite{bai2022transfusion, li2022bevdepth, li2022bevformer, yin2021center}. \par 
\begin{table}[t]
    \captionsetup{font=small, belowskip=-11pt}
    \footnotesize
    \centering
    \setlength{\tabcolsep}{7.3pt}
    \begin{tabular}{l|cc|c|cc}
        \toprule
        \textit{Model} 
        & \textit{LSS++} 
        & \textit{DS} 
        & GFLOPS
        & \cellcolor[HTML]{ab9ac0}\textit{mAP}$\uparrow$ 
        & \cellcolor[HTML]{a5eb8d}\textit{NDS}$\uparrow$ \\
        \midrule
         \multirow{4}{*}{BEVDepth$^\dagger$} 
         & \xm & \xm & 298 & 32.4 & 44.9 \\
         & \xm & \ch & 298 & 33.1 & 44.9 \\
        & \ch & \xm & 316 & 34.9 & 47.0 \\
        & \ch & \ch & 316 & 35.9 & 47.2 \\
        \midrule
        \textbf{X$^3$KD} (Ours) & \ch & \ch & 316 & \textbf{39.0} & \textbf{50.5} \\
        \bottomrule
    \end{tabular}\vspace{-5pt}
    \caption{\textbf{Analysis of BEVDepth$^\dagger$} (re-implementation of~\cite{li2022bevdepth}): We compare the architectural improvement of a larger Lift-Splat-Shoot (LSS++) transform to using depth supervision (DS).}
    \label{tab:motivation-analysis}
\end{table}
While LiDAR scanners may not be available in commercially deployed vehicle fleets, they are typically available in training data collection vehicles to facilitate 3D annotation. Therefore, LiDAR data is privileged; it is often available during training but not during inference. The recently introduced BEVDepth~\cite{li2022bevdepth} approach pioneers using accurate 3D information from LiDAR data at training time to improve multi-camera 3DOD, see~Fig.~\ref{fig:concept} (top part). Specifically, it proposed an improved Lift-Splat-Shoot PV-to-BEV transform (LSS++) and depth supervision (DS) by projected LiDAR points, which we analyze in Table~\ref{tab:motivation-analysis}. We observe that the LSS++ architecture yields significant improvements, though depth supervision seems to have less effect. This motivates us to find additional types of supervision to transfer accurate 3D information from LiDAR point clouds to multi-camera 3DOD. To this end, we propose cross-modal knowledge distillation (KD) to not only use LiDAR \textit{data} but a high-performing LiDAR-based 3DOD \textit{model}, as in~Fig.~\ref{fig:concept} (middle part). To provide an overview of the effectiveness of cross-modal KD at various multi-camera 3DOD network stages, we present three distillation techniques: feature distillation (X-FD) and adversarial training (X-AT) to improve the feature representation by the intermediate information contained in the LiDAR 3DOD model as well as output distillation (X-OD) to enhance output-stage supervision.
\par 
For optimal camera-based 3DOD, extracting useful PV features before the view transformation to BEV is equally essential. However, gradient-based optimization through an ambiguous view transformation can induce non-optimal supervision signals. Recent work proposes pre-training the PV feature extractor on instance segmentation to improve the extracted features~\cite{xie2022m2bev}. Nevertheless, neural networks are subject to catastrophic forgetting~\cite{kirkpatrick2016overcoming} such that knowledge from pre-training will continuously degrade if not retained by supervision. Therefore, we propose cross-task instance segmentation distillation (X-IS) from a pre-trained instance segmentation teacher into a multi-camera 3DOD model, see~Fig.~\ref{fig:concept} (bottom part). As shown in Table~\ref{tab:motivation-analysis}, our X$^3$KD framework significantly improves upon BEVDepth without additional complexity during inference.
\par
To summarize, our main contributions are as follows: 
\vspace{-5pt}
\begin{itemize}
\setlength\itemsep{-3.0pt}
\item We propose X$^3$KD, a KD framework across modalities, tasks, and stages for multi-camera 3DOD.
\item Specifically, we introduce cross-modal KD from a strong LiDAR-based 3DOD teacher to the multi-camera 3DOD student, which is applied at multiple network stages in bird's eye view, \ie, feature-stage (X-FD and X-AT) and output-stage (X-OD).
\item Further, we present cross-task instance segmentation distillation (X-IS) at the PV feature extraction stage.
\item X$^3$KD outperforms previous approaches for multi-camera 3DOD on the nuScenes and Waymo datasets. 
\item We transfer X$^3$KD to RADAR-based 3DOD and train X$^3$KD only through KD without using ground truth.
\item Our extensive ablation studies on nuScenes and Waymo provide a comprehensive evaluation of KD at different network stages for multi-camera 3DOD.
\end{itemize}

\section{Related Work}
\label{sec:related_work}
\vspace{-2pt}

\textbf{Multi-View Camera-Based 3D Object Detection}:
Current multi-view 3D object detectors can be divided into two main streams: First, DETR3D and succeeding works \cite{liu2022petr, li2022bevformer, liu2022petrv2, wang2022detr3d, zhou2022crossview} project a sparse set of learnable 3D queries/priors onto 2D image features with subsequent sampling and an end-to-end 3D bounding box regression. Second, LSS and following works \cite{philion2020lift, huang2021bevdet, li2022bevdepth} employ a view transformation consisting of a depth prediction, a point cloud reconstruction, and a voxel pooling to project points to BEV. 3D bounding boxes are predicted from these BEV features. While such works focus on improving the network architecture and view transformation, we focus on better model optimization. In this direction, M$^2$BEV~\cite{xie2022m2bev} proposed instance segmentation pre-training of the PV feature extraction. We propose cross-task instance segmentation distillation to retain this knowledge during 3DOD training.
\par 
Most current state-of-the-art works focus on incorporating temporal information either through different kinds of feature-level aggregation~\cite{huang2022bevdet4d, li2022bevformer, li2022bevdepth, liu2022petrv2} or by improving depth estimation by temporal stereo approaches~\cite{li2022bevstereo, wang2022sts}. While the usual setting considers data from 2 time steps, recently proposed SOLOFusion \cite{Park2022timewt} separately models long-range and short-range temporal dependencies in input data from 16 time steps. Our work focuses on a different direction, \ie, we try to optimally exploit the information contained in LiDAR point clouds. In this direction, BEVDepth~\cite{li2022bevdepth} and succeeding works~\cite{li2022bevstereo, Park2022timewt} supervise the depth estimation with projected LiDAR points. We explore this path further by using cross-modal knowledge distillation (KD) from a LiDAR-based 3DOD teacher.
\par 
\textbf{Multi-Modal 3D Object Detection}:
Recently, there has been a trend to fuse different sensor modalities, especially camera and LiDAR, with the idea of combining modality-specific useful information, hence improving the final 3DOD performance \cite{bai2022transfusion, liu2022bevfusion, li2022deepfusion, yang2022deepinteraction, jiao2022msmdfusion, fusionpainting}. 
Existing 3DOD methods mostly perform multi-modal fusion at one of the three stages: First, various approaches \cite{vora2020pointpainting, wang2021pointaugmenting, fusionpainting} propose to decorate/augment the raw LiDAR points with image features. Second, intermediate feature fusion of the modalities in a shared representation space, such as the BEV space, has been explored~\cite{chen2022autoalign, liu2022bevfusion, li2022deepfusion, jiao2022msmdfusion, yang2022deepinteraction}. Third, proposal-based fusion methods~\cite{bai2022transfusion, ku2018joint, chen2022futr3d} keep the feature extraction of different modalities independent and aggregate multi-modal features via proposals or queries in the 3DOD prediction head. 
While these approaches require both sensors to be available during inference, our X$^3$KD approach requires only camera sensors during inference. We also apply our KD approach to less frequently explored RADAR- and camera-RADAR fusion-based models.
\par 
\textbf{Knowledge Distillation for 3D Object Detection:}
Employing the KD technique from~\cite{hinton2015distilling} some recent works have explored KD for 3DOD~\cite{chong2022monodistill, liu2022multi, wei2022lidar, zhang2022pointdistiller}. Most works focus on LiDAR-based 3DOD settings and propose methods to improve performance or efficiency~\cite{yang2022towards, zhang2022pointdistiller} or solve problems that are specific to point clouds, such as KD into sparser point clouds~\cite{wei2022lidar, zheng2022boosting}. Some initial works have also proposed concepts for cross-modal KD in 3D semantic segmentation~\cite{liu2021distillation} or simple single or stereo camera-based 3DOD models~\cite{chong2022monodistill, guo2021liga, hong2022cross, liu2022multi, zhou2022sgm3d}. However, current research focus has shifted to more general multi-camera settings, where up to our knowledge, we are the first to investigate KD across modalities, tasks, and stages comprehensively.

\section{Proposed X$^3$KD Framework}
\label{sec:method}
\vspace{-3pt}

We first define our considered problem and baseline in Sec.~\ref{sec:method_problem}. Next, we give an overview on X$^3$KD in Sec.~\ref{sec:method_overview} presenting specific advancements in Secs.~\ref{sec:method_cross_modal} and \ref{sec:method_cross_task}.
\subsection{Problem Formulation and Baseline Method}
\vspace{-3pt}
\label{sec:method_problem}

\textbf{Problem Definition}:
We aim at developing a 3DOD model with camera images $\bm{x}\in\mathbb{R}^{N^\text{cam}\times H^\text{cam}\times W^\text{cam}\times 3}$ as input, where $N^\text{cam}$, $H^\text{cam}$, and $W^\text{cam}$ represent the number of images, image height, and image width, respectively, and $N^{\text{bbox}}$ 3D bounding boxes $\overline{\bm{b}} = \big\lbrace \left(\overline{\bm{b}}_n^{\text{reg}}, \overline{b}_n^{\text{cls}}\right) n\in\big\lbrace 1,\ldots, N^{\text{bbox}}\big\rbrace\big\rbrace$ as output.
Each bounding box is represented by regression parameters $\overline{\bm{b}}_n^{\text{reg}}\in\mathbb{R}^9$ (three, three, two, and one for the center, spatial extent, velocity, and yaw angle, respectively), and a classification label $\overline{b}_n^{\text{cls}}\in\mathcal{S}$ from the set of $|\mathcal{S}|$ classes $\mathcal{S}=\big\lbrace 1, \ldots, |\mathcal{S}| \big\rbrace$.
During training, not only are camera images available, but we can also make use of a 3D LiDAR point cloud $\bm{l}\in\mathbb{R}^{P\times 5}$ with $P$ points, each one containing the 3D position, intensity, and ring index. The point cloud $\bm{l}$ is not available during inference.\par
\textbf{Baseline Model}:
We build upon the recently published state-of-the-art method BEVDepth~\cite{li2022bevdepth}, whose setup is depicted in the blue box of Fig.~\ref{fig:method_overview}. First, all images are processed by a PV feature extractor, yielding features $\bm{f}^\text{PV}\in\mathbb{R}^{N^\text{cam}\times H^\text{PV}\times W^\text{PV}\times C^\text{PV}}$ in PV with spatial extent $H^\text{PV}\times W^\text{PV}$ and number of channels $C^\text{PV}$. Afterwards, the features are passed through the Lift-Splat-Shoot transform~\cite{philion2020lift}, which predicts discretized depth values $\hat{\bm{d}}$, transforms pixels corresponding to $\bm{f}^\text{PV}$ into a point cloud representation and obtains BEV features $\bm{f}^\text{BEV}\in\mathbb{R}^{H^\text{BEV}\times W^\text{BEV}\times C^\text{BEV}}$ via voxel pooling. BEV features are further processed by an encoder-decoder network as in~\cite{li2022bevdepth}, yielding refined features $\bm{f}^\text{REF}\in\mathbb{R}^{H^\text{BEV}\times W^\text{BEV}\times C^\text{REF}}$. Finally, the CenterPoint prediction head~\cite{yin2021center}, predicts dense object probability scores $\hat{\bm{b}}^{\text{cls}}\in\mathbb{I}^{H^\text{BEV}\times W^\text{BEV}\times |\mathcal{S}|}$ for each class as well as corresponding regression parameters $\hat{\bm{b}}^{\text{reg}}\in\mathbb{R}^{H^\text{BEV}\times W^\text{BEV}\times 9}$. The final bounding box predictions $\overline{\bm{b}}$ are generated by non-learned decoding of these dense representations~\cite {yin2021center}. 
\par
\textbf{Baseline Training}: 
The baseline is trained by optimizing the 3D bounding box losses $\mathcal{L}^{\text{CPoint}}$ from Centerpoint~\cite{yin2021center} as well as the depth loss $\mathcal{L}^{\text{depth}}$ from~\cite{li2022bevdepth}, yielding 
\begin{equation}
	\mathcal{L}^{\text{GT}} = \mathcal{L}^{\text{depth}}(\hat{\bm{d}}, \bm{d}) + \mathcal{L}^{\text{CPoint}}(\hat{\bm{b}}^{\text{cls}}, \hat{\bm{b}}^{\text{reg}}, \bm{b}),
    \label{eq:3dod-supervised-training}
\end{equation}
where $\bm{d}$ is the depth ground truth generated from projected LiDAR points and $\bm{b}$ is the set of ground truth bounding boxes. For more details, we refer to the supplementary.
\subsection{X$^3$KD Overview}
\vspace{-3pt}
\label{sec:method_overview}

\begin{figure*}
	\captionsetup{font=small, belowskip=-14pt}
    \centering
	\includegraphics[width=0.95\textwidth]{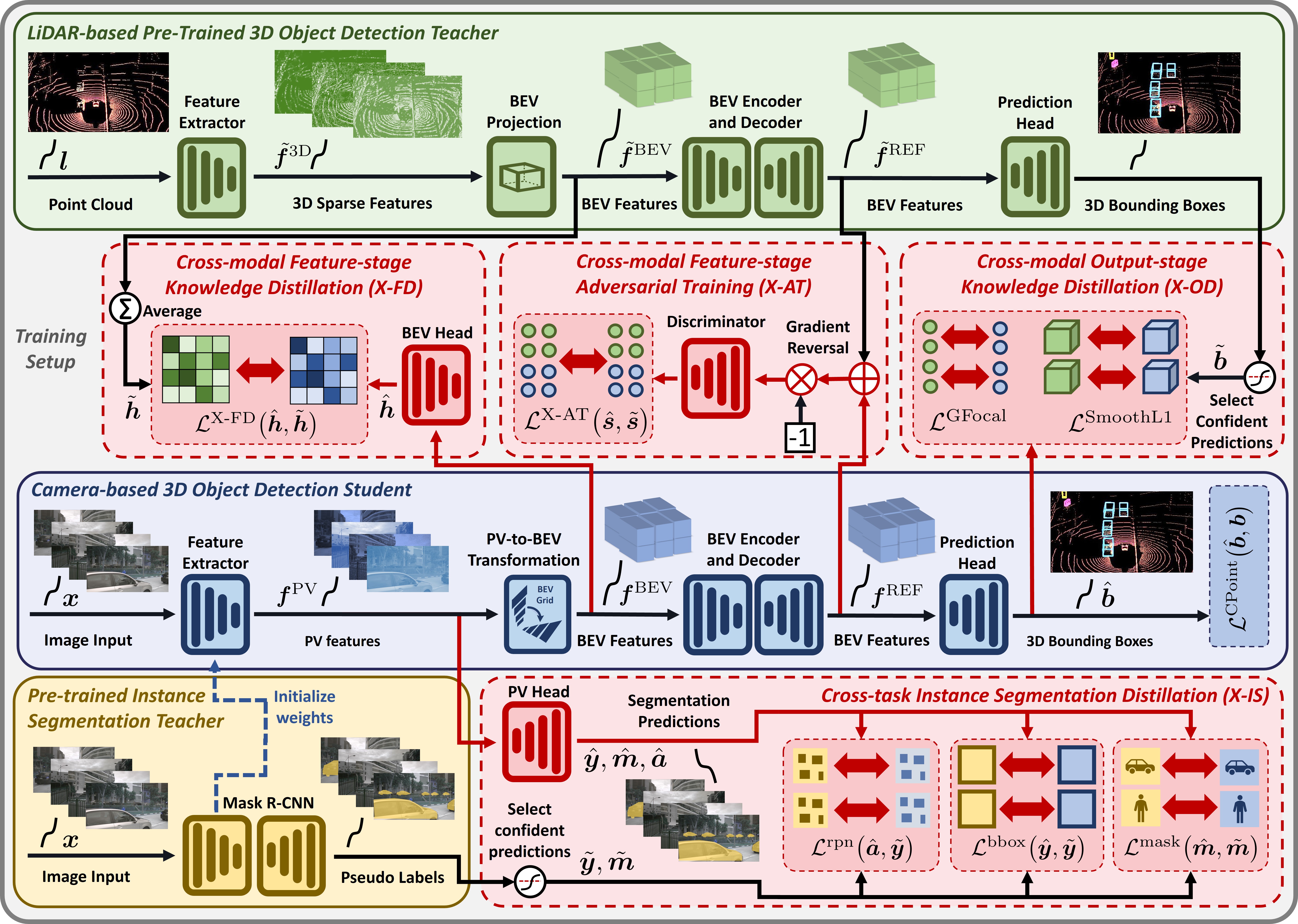}
    \vspace{-6pt}
    \caption{\textbf{We present X$^3$KD, a knowledge distillation (KD) framework for multi-camera 3DOD.} We employ an inference setup (middle blue box) relying only on multi-camera image input (LiDAR point cloud in the output is just shown for visualization). During training, we apply KD across several network stages (red arrows originating from the blue box): In perspective-view (PV) feature extraction, we apply cross-task instance segmentation distillation (X-IS) from an instance segmentation teacher (yellow box). In the bird's eye view (BEV), we apply cross-modal feature distillation (X-FD), adversarial training (X-AT), and output distillation (X-OD) from a LiDAR-based 3DOD teacher (green box). X$^3$KD significantly enhances the multi-camera 3DOD without inducing extra complexity during inference.}
    \label{fig:method_overview}
\end{figure*} 

Our X$^3$KD framework (Fig.~\ref{fig:method_overview}) improves the performance of a multi-camera 3DOD model without introducing additional complexity during inference. Hence, our model's inference setup is equal to the one of our baseline. During training, however, we explore multiple knowledge distillation (KD) strategies across modalities, tasks, and stages.
\par
\textbf{X$^3$KD Loss}: First, we employ a pre-trained LiDAR-based 3DOD model, as shown in~Fig.~\ref{fig:method_overview} (top part). We propose three losses for distilling knowledge across different stages into the camera-based 3DOD: An output-stage distillation (X-OD) loss $\mathcal{L}^{\text{X-OD}}$ between the outputs of the camera and LiDAR models, a feature-stage distillation (X-FD) scheme and a corresponding loss $\mathcal{L}^{\text{X-FD}}$ to guide the focus of the BEV features after the view transformation, and a feature-stage adversarial training (X-AT) with a loss $\mathcal{L}^{\text{X-AT}}$ between the camera and LiDAR model features to encourage their feature similarity. Second, we use an instance segmentation network, cf.~Fig.~\ref{fig:method_overview} (bottom part). We propose cross-task instance segmentation distillation (X-IS) by imposing a loss $\mathcal{L}^{\text{X-IS}}$ between the output of an additional PV instance segmentation head and teacher-generated pseudo labels. Our total loss for X$^3$KD is then given by:
\begin{equation}
    \mathcal{L}^{\mathrm{X^3KD}} \!\!=\!\! \sum_{i\in\mathcal{I}} \lambda^{i} \mathcal{L}^{i}, \mathcal{I}\!=\!\left\lbrace\fontsize{9}{9}\selectfont\text{GT}, \text{X-OD}, \text{X-FD}, \text{X-AT}, \text{X-IS}\right\rbrace
    \label{eq:x3kd-loss}
\end{equation}

\subsection{Cross-modal Knowledge Distillation} 
\vspace{-3pt}
\label{sec:method_cross_modal}

The current superiority of LiDAR-based 3DOD over multi-camera 3DOD can be attributed to the ambiguous view transformation in multi-camera models, which may place features at the wrong position in the final representation (\eg, a BEV grid). Meanwhile, LiDAR-based models operate on a 3D point cloud, which can easily be projected onto any view representation. Thereby, the extracted features preserve 3D information. Our cross-modal KD components transfer this knowledge to the multi-camera 3DOD model across different network stages, cf.~Fig.~\ref{fig:method_overview} (top part). 
\par 
\textbf{LiDAR-based 3DOD Model Architecture}: Our LiDAR-based 3DOD model is mainly based on CenterPoint~\cite{yin2021center}. First, the point cloud $\bm{l}\in\mathbb{R}^{P\times 5}$ is processed by the Sparse Encoder from SECOND~\cite{yan2018second}, yielding 3D sparse features $\tilde{\bm{f}}^{\text{3D}}\in \mathbb{R}^{H^\text{BEV}\times W^\text{BEV}\times \tilde{D}^\text{3D}\times \tilde{C}^\text{3D}}$ with volumetric extent $H^\text{BEV}\times W^\text{BEV}\times \tilde{D}^\text{3D}$ and number of channels $\tilde{C}^\text{3D}$. Then, the features are projected onto the same BEV plane as in the camera-based 3DOD model, yielding BEV features $\tilde{\bm{f}}^{\text{BEV}}\in \mathbb{R}^{H^\text{BEV}\times W^\text{BEV}\times \tilde{C}^\text{BEV}}$ with $\tilde{C}^\text{BEV} = \tilde{D}^\text{3D} \cdot \tilde{C}^\text{3D}$. These are further processed by an encoder-decoder network, yielding refined BEV features $\tilde{\bm{f}}^{\text{REF}}\in \mathbb{R}^{H^\text{BEV}\times W^\text{BEV}\times \tilde{C}^\text{REF}}$. Finally, the features are passed through a prediction head, yielding probability score maps $\tilde{\bm{b}}^{\text{cls}}\in\mathbb{I}^{H^\text{BEV}\times W^\text{BEV}\times |\mathcal{S}|}$ and regression maps $\tilde{\bm{b}}^{\text{reg}}\in\mathbb{R}^{H^\text{BEV}\times W^\text{BEV}\times 9}$ analogous to the outputs $\hat{\bm{b}}^{\text{cls}}$ and $\hat{\bm{b}}^{\text{reg}}$ of the multi-camera 3DOD model. 
\par
\textbf{Output-stage Distillation (X-OD)}:
Following many approaches in KD~\cite{dai2021general, hinton2015distilling, yang2022focal}, we distill knowledge at the output stage by imposing losses between the teacher's outputs $\tilde{\bm{b}}^{\text{cls}}$ and $\tilde{\bm{b}}^{\text{reg}}$ and the student's outputs $\hat{\bm{b}}^{\text{cls}}$ and $\hat{\bm{b}}^{\text{reg}}$. Specifically, we impose a Gaussian focal loss $\mathcal{L}^{\text{GFocal}}$~\cite{law2018cornernet} between $\hat{\bm{b}}^{\text{cls}}$ and $\tilde{\bm{b}}^{\text{cls}}$ to put more weight on rare classes and compensate for the class imbalance. As this loss only considers pseudo labels as a positive sample if they are exactly $1$, we select high-confidence teacher output probabilities $\tilde{\bm{b}}^{\text{cls}}$, \ie, probability values over a threshold $\alpha^{\text{3D-bbox}}$, and set them to $1$. Further, the regression output of the student $\hat{\bm{b}}^{\text{reg}}$ is supervised by the corresponding output $\tilde{\bm{b}}^{\text{reg}}$ of the teacher by imposing a Smooth L1 loss $\mathcal{L}^{\text{SmoothL1}}$~\cite{girshick2015fast}. Finally, we propose to weigh the regression loss by the teacher's pixel-wise averaged output probabilities $\langle \tilde{\bm{b}}_s^{\text{cls}} \rangle = \frac{1}{|\mathcal{S}|}\sum_{s\in\mathcal{S}} \tilde{\bm{b}}_s^{\text{cls}} \in\mathbb{R}^{H^\text{BEV}\times W^\text{BEV}}$ to weigh regions which likely contain objects higher than the background. Overall, X-OD is defined as:
\begin{equation}
	\mathcal{L}^{\text{X-OD}}\big( \hat{\bm{b}}, \tilde{\bm{b}}\big)\! =\! \mathcal{L}^{\text{GFocal}}\big( \hat{\bm{b}}^{\text{cls}}, \tilde{\bm{b}}^{\text{cls}}\big)\! + \! \mathcal{L}^{\text{SmoothL1}}\big( \hat{\bm{b}}^{\text{reg}}, \tilde{\bm{b}}^{\text{reg}}\big)
 \label{eq:x-od}
\end{equation}
\par
\begin{figure}
    \captionsetup{font=small, belowskip=-14pt}
    \centering
    \vspace{-5pt}
    \includegraphics[width=0.95\linewidth]{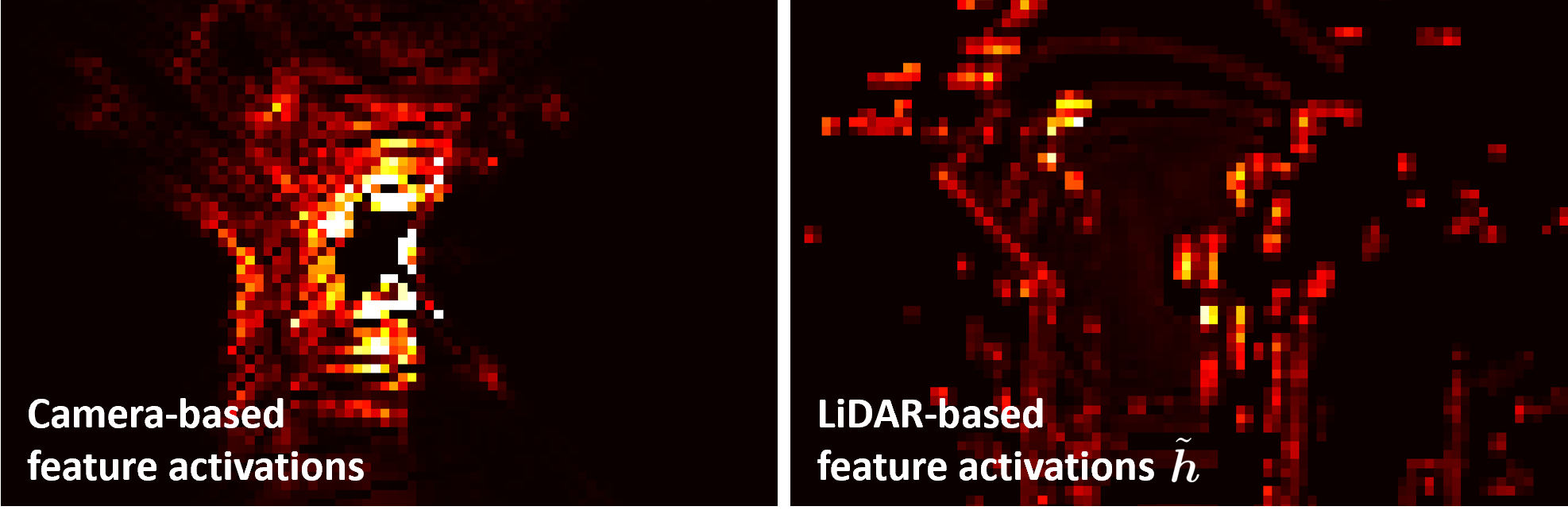}
    \vspace{-6pt}
    \caption{\textbf{Mean feature activations} from the camera-based student after the view transformation (left) and the LiDAR-based teacher (right) exhibit structural dissimilarity.}
    \label{fig:features}
\end{figure}
\textbf{Feature-stage Distillation (X-FD)}:
Our X-FD component exploits the precise and sparse nature of features extracted from LiDAR point clouds, which precisely encode locations of relevant objects for 3DOD. Thereby, the mean sparse feature activation $\tilde{\bm{h}}$, cf.~Fig.~\ref{fig:features} (right), provides a good initial estimate for the potential location of objects. While it would be natural to impose similarity losses between BEV features from the camera and LiDAR models, these features are structurally quite different (cf.~Fig.~\ref{fig:features}), such that our attempts to impose such losses did lead to unstable training behavior. Therefore, we add a small BEV decoder to the multi-camera model, which outputs a prediction $\hat{\bm{h}}$ for the mean sparse feature activations from the LiDAR teacher $\tilde{\bm{h}}$. The X-FD loss $\mathcal{L}^{\text{X-FD}}$ is then given as:
\begin{equation}
	\mathcal{L}^{\text{X-FD}} = \mathrm{L1}\big( \hat{\bm{h}}, \tilde{\bm{h}} \big)
\end{equation}
\par 
\textbf{Feature-stage Adversarial Training (X-AT)}:
We further propose X-AT to encourage a more global feature similarity between the refined features $\bm{f}^{\text{REF}}$ and $\tilde{\bm{f}}^{\text{REF}}$ from both modalities in BEV space. Due to the structural dissimilarity of features from both modalities directly after the BEV projection (Fig.~\ref{fig:features}), we apply the adversarial training on the refined features $\tilde{\bm{f}}^{\text{REF}}$ and $\hat{\bm{f}}^{\text{REF}}$. We pass these cross-modal features through a gradient reversal layer and a patch-based discriminator network~\cite{isola2017image}, which outputs two modality-specific probabilities. The discriminator is optimized to classify the features by modality using a binary cross-entropy loss $\mathcal{L}^{\text{X-AT}}$ between the output probabilities $\hat{\bm{s}}$ and the ground truth modality labels $\bm{s}$:
\begin{equation}
	\mathcal{L}^{\text{X-AT}} = \mathrm{BCE}\left( \hat{\bm{s}}, \bm{s} \right)
\end{equation}
We then encourage modality-agnostic features in the multi-camera 3DOD model through gradient reversal.
\subsection{Cross-task Knowledge Distillation} 
\vspace{-3pt}
\label{sec:method_cross_task}

Learning a good feature representation in PV is difficult when all supervision signals are backpropagated through an ambiguous view transformation. As a possible solution, M$^2$BEV~\cite{xie2022m2bev} proposes instance segmentation (IS) pre-training. However, deep neural networks exhibit catastrophic forgetting such that this initial knowledge is not necessarily preserved during 3DOD training. Therefore, we propose cross-task instance segmentation distillation (X-IS) to preserve the knowledge contained in the PV features continuously. Specifically, we use the outputs of a pre-trained instance segmentation network as pseudo labels to optimize an additional PV instance segmentation head, cf.~Fig.~\ref{fig:method_overview}.
\par 
\textbf{Pseudo Label Generation}: 
In this work, we use the well-established Mask R-CNN architecture~\cite{he2017mask} as a teacher; see Fig.~\ref{fig:method_overview} (bottom left). We use its original architecture, consisting of a feature extractor, a feature pyramid network (FPN), a region proposal network (RPN), and a region of interest (ROI) head, including a mask branch. As output, we obtain $N^{\text{IS}}$ bounding boxes $\tilde{\bm{y}} = \big\lbrace \big( \tilde{\bm{y}}_n^{\text{bbox}}, \tilde{y}_n^{\text{cls}}, \tilde{y}_n^{\text{score}}\big), n\in\big\lbrace 1,\ldots, N^{\text{IS}}\big\rbrace \big\rbrace$ with four parameters for bounding box center and spatial extent $\tilde{\bm{y}}_n^{\text{bbox}}\in\mathbb{R}^4$, a classification result $\tilde{y}_n^{\text{cls}}\in\mathcal{S}^{\text{IS}}$ from the set of IS classes $\mathcal{S}^{\text{IS}}$, and an objectness score $\tilde{y}_n^{\text{score}}\in\mathbb{I}$ with $\mathbb{I} = \left[0,1\right]$. Additionally, we obtain corresponding object masks $\tilde{\bm{m}} = \big\lbrace \tilde{\bm{m}}_n, n\in\big\lbrace 1,\ldots, N^{\text{IS}}\big\rbrace \big\rbrace$ with single masks $\tilde{\bm{m}}_n\in\left\lbrace 0, 1\right\rbrace^{H^{\text{mask}}_n\times W^{\text{mask}}_n}$ and spatial resolution $H^{\text{mask}}_n\times W^{\text{mask}}_n$. We select all samples with a score $\tilde{y}_n^{\text{score}} > \alpha^{\text{2D-bbox}}$ as pseudo labels.\par 
\begin{table*}[t]
\captionsetup{font=small, belowskip=-13pt}
\footnotesize
\centering
\def\arraystretch{0.95}%
\setlength{\tabcolsep}{8.55pt}
    \centering
    \begin{tabular}{c|l|c|c|ccccc|cc}
    \toprule
    \textit{Set} & \textit{Model} & 
    \textit{Backbone} & 
    \textit{Resolution} & 
    \cellcolor[HTML]{96bbce}\textit{mATE}$\downarrow$ & 
    \cellcolor[HTML]{fb9a99}\textit{mASE}$\downarrow$ & 
    \cellcolor[HTML]{fc4538}\textit{mAOE}$\downarrow$ & 
    \cellcolor[HTML]{fdbf6f}\textit{mAVE}$\downarrow$ & 
    \cellcolor[HTML]{ff7f00}\textit{mAAE}$\downarrow$ & 
    \cellcolor[HTML]{ab9ac0}\textit{mAP}$\uparrow$ & 
    \cellcolor[HTML]{a5eb8d}\textit{NDS}$\uparrow$ \\
    \midrule
    \multirow{7}{*}{\rotatebox{90}{Validation}}
     & BEVDet~\cite{huang2021bevdet} & \multirow{6}{*}{ResNet-50} & \multirow{6}{*}{$256\times 704$} & 0.725 & 0.279 & 0.589 & 0.860 & 0.245 & 29.8 & 37.9 \\       
    & BEVDet4D~\cite{xie2022m2bev}     & & & 0.703 & 0.278 & 0.495 & 0.354 & 0.206 & 32.2 & 45.7 \\ 
    & BEVDepth~\cite{li2022bevdepth}   & & & 0.629 & \textbf{0.267} & 0.479 & 0.428 & 0.198 & 35.1 & 47.5 \\
    & BEVDepth$^\dagger$               & & & 0.636 & 0.272 & 0.493 & 0.499 & 0.198 & 35.9 & 47.2 \\
    & STS$^*$~\cite{wang2022sts}           & & & 0.601 & 0.275 & 0.450 & 0.446 & 0.212 & 37.7 & 48.9 \\
    & BEVStereo$^*$~\cite{li2022bevstereo} & & & \textbf{0.598} & 0.270 & \textbf{0.438} & 0.367 & \textbf{0.190} & 37.2 & 50.0 \\
    & \CC{20} \textbf{X$^{3}$KD}$_{\mathrm{all}}$ & \CC{20} ResNet-50 & \CC{20} $256\times 704$ & \CC{20} 0.615 \CC{20} & \CC{20} 0.269 & \CC{20} 0.471 & \CC{20} \textbf{0.345} & \CC{20} 0.203 & \CC{20} \textbf{39.0} & \CC{20} \textbf{50.5} \\
    \midrule 
    \multirow{5}{*}{\rotatebox{90}{Validation}} & PETR~\cite{liu2022petr}                     & \multirow{4}{*}{ResNet-101} & \multirow{4}{*}{$512\times 1408$} & 0.710 & 0.270 & 0.490 & 0.885 & 0.224 & 35.7 & 42.1 \\
    & BEVDepth$^\dagger$             & &  &  0.579 &  0.265 &  0.387 &  0.364 &  0.194 &  40.9 &  53.1 \\
    & BEVDepth~\cite{li2022bevdepth} & & & 0.565 & 0.266 & 0.358 & 0.331 & \textbf{0.190} & 41.2 & 53.5 \\
    & STS$^*$~\cite{wang2022sts}         & & & \textbf{0.525} & 0.262 & 0.380 & 0.369 & 0.204 & 43.1 & 54.2 \\
    & \CC{20} \textbf{X$^{3}$KD}$_{\mathrm{all}}$ & \CC{20} ResNet-101 & \CC{20} $512\times 1408$ & \CC{20} 0.552 & \CC{20} \textbf{0.257} & \CC{20} \textbf{0.338} & \CC{20} \textbf{0.328} & \CC{20} 0.199 & \CC{20} \textbf{44.8} & \CC{20} \textbf{55.3} \\
    \midrule 
    \multirow{5}{*}{\rotatebox{90}{Validation}} & DETR3D~\cite{wang2022detr3d}                   & \multirow{3}{*}{ResNet-101} & \multirow{3}{*}{$900\times 1600$} & 0.716 & 0.268 & 0.379 & 0.842 & 0.200 & 34.9 & 43.4 \\
    & BEVFormer~\cite{li2022bevformer} & & & 0.673 & 0.274 & 0.372 & 0.394 & 0.198 & 41.6 & 51.7 \\
    & PolarFormer~\cite{jiang2022polarformer}              & & & 0.648 & 0.270 & 0.348 & 0.409 & 0.201 & 43.2 & 52.8 \\
    &  BEVDepth$^\dagger$               & ResNet-101 &  $640\times 1600$ &  0.571 &  0.260 &  0.379 &  0.374 &  \textbf{0.196} &  42.8 &  53.6 \\
    & \CC{20} \textbf{X$^{3}$KD}$_{\mathrm{all}}$ & \CC{20} ResNet-101 & \CC{20} $640\times 1600$ & \CC{20} \textbf{0.539} & \CC{20} \textbf{0.255} & \CC{20} \textbf{0.320} & \CC{20} \textbf{0.324} & \CC{20} \textbf{0.196} & \CC{20} \textbf{46.1} & \CC{20} \textbf{56.7} \\
    \midrule   
    \multirow{4}{*}{\rotatebox{90}{Test}} & BEVFormer~\cite{li2022bevformer} & & & 0.631 & 0.257 & 0.405 & 0.435 & 0.143 & 44.5 & 53.5 \\
    &  BEVDepth$^\dagger$ &  ResNet-101 &  $640\times 1600$ &  0.533 &  0.254 &  0.443 &  0.404 &  \textbf{0.129} &  43.1 &  53.9 \\
    & PolarFormer~\cite{jiang2022polarformer}              & & & 0.610 & 0.258 & \textbf{0.391} & 0.458 & \textbf{0.129} & \textbf{45.6} & 54.3 \\
    & \CC{20} \textbf{X$^{3}$KD}$_{\mathrm{all}}$  & \CC{20} ResNet-101 & \CC{20} $640\times 1600$ & \CC{20} \textbf{0.506} & \CC{20} \textbf{0.253} & \CC{20} 0.414 & \CC{20} \textbf{0.366} & \CC{20} 0.131 & \CC{20} \textbf{45.6} & \CC{20} \textbf{56.1} \\
    \bottomrule
    \end{tabular}
    \vspace{-6pt}
    \caption{\textbf{Performance comparison on the nuScenes dataset}: We ensure comparability regarding backbone and image resolution. Baseline results are cited except for BEVDepth$^\dagger$ which we reproduced in our framework; $^*$ indicates recent ArXiv works; best numbers in boldface.}
    \label{tab:sota-nuscenes}
\end{table*}

\textbf{X-IS Loss Computation}: 
The teacher-generated pseudo labels are used to supervise an additional PV instance segmentation head, cf~Fig.~\ref{fig:method_overview} (bottom right), which uses the same RPN and ROI head architectures as the teacher. The RPN head outputs region proposals $\hat{\bm{a}} = \big( \hat{\bm{a}}^{\text{cls}}, \hat{\bm{a}}^{\text{reg}}\big)$ with foreground/background scores $\hat{\bm{a}}^{\text{cls}}\in\mathbb{I}^{H^\text{PV}\times W^\text{PV}\times 2K}$ and regression parameters $\hat{\bm{a}}^{\text{reg}}\in\mathbb{R}^{H^\text{PV}\times W^\text{PV}\times 4K}$ relative to each of the $K$ anchors. Our RPN loss $\mathcal{L}^{\text{rpn}}$ is then comprised of an assignment strategy between pseudo GT and PV head outputs as detailed in~\cite{ren2017faster} and subsequent application of BCE and L1 differences for optimizing $\hat{\bm{a}}^{\text{cls}}$ and $\hat{\bm{a}}^{\text{reg}}$, respectively. The $N^{\text{RPN}}$ region proposals with the highest foreground scores are subsequently passed through the ROI head, which outputs refined bounding boxes $\hat{\bm{y}}=\big\lbrace \big( \hat{\bm{y}}_n^{\text{bbox}}, \hat{y}_n^{\text{cls}}\big), n\in\big\lbrace 1,\ldots, N^{\text{RPN}}\big\rbrace \big\rbrace$ with class probabilities $\hat{\bm{y}}_n^{\text{cls}}\in\mathbb{I}^{|\mathcal{S}^{\text{IS}}|}$, four bounding box regression parameters $\hat{\bm{y}}_n^{\text{bbox}}\in\mathbb{R}^4$ as well as class-specific mask probabilities $\hat{\bm{m}} = \big\lbrace \hat{\bm{m}}_n, n\in\big\lbrace 1,\ldots, N^{\text{IS}}\big\rbrace \big\rbrace$ with single masks $\hat{\bm{m}}_n\in \mathbb{I}^{H^\text{mask}\times W^\text{mask}\times |\mathcal{S}^{\text{IS}}|}$. Our bounding box loss $\mathcal{L}^{\text{bbox}}$ is comprised of an assignment strategy between ground truth $\tilde{\bm{y}}$ and prediction $\hat{\bm{y}}$ and subsequent application of L1 difference between $\hat{\bm{y}}_n^{\text{bbox}}$ and $\tilde{\bm{y}}_n^{\text{bbox}}$ as well as cross-entropy (CE) difference between $\hat{\bm{y}}_n^{\text{cls}}$ and one-hot encoded $\tilde{\bm{y}}_n^{\text{cls}}$. For computing the mask loss $\mathcal{L}^{\text{mask}}$, we apply a binary cross entropy (BCE) difference between ground truth $\tilde{\bm{m}}$ and prediction $\hat{\bm{m}}$, selecting only the output corresponding to the ground truth mask's class. More details can be found in~\cite{he2017mask}. Overall, our X-IS loss $\mathcal{L}^{\text{X-IS}}$ can be written as:
\begin{equation}
    \mathcal{L}^{\text{X-IS}} = \mathcal{L}^{\text{rpn}}\left(\hat{\bm{a}}, \tilde{\bm{y}}\right) + \mathcal{L}^{\text{bbox}}\left(\hat{\bm{y}}, \tilde{\bm{y}}\right) + \mathcal{L}^{\text{mask}}\left(\hat{\bm{m}}, \tilde{\bm{m}}\right).
\end{equation}
\section{Experiments}
\label{sec:experiments}
\vspace{-3pt}

We first provide our experimental setup (Sec.~\ref{sec:exp-setup}) and a state-of-the-art comparison (Sec.~\ref{sec:sota-comp}). Next, we verify and analyze our method's components in Secs.~\ref{sec:ablation} and \ref{sec:analysis}. Last, we evaluate RADAR-based models (Sec.~\ref{sec:gen-to-radar}).
\begin{table}[t]
	\captionsetup{font=small, belowskip=-12pt}
    \centering
    \footnotesize
    \def\arraystretch{0.97}%
    \setlength{\tabcolsep}{4.8pt}
    \begin{tabular}{l|c|ccc|c}
        \toprule
        \multirow{2}{*}{\textit{Model}} & \textit{LET-3D-AP}$\uparrow$ & \multicolumn{4}{c}{\textit{LET-3D-APL$\uparrow$}} \\
        & \textit{All} & \textit{Vehicle} & \textit{Pedestrian} & \textit{Cyclist} & \textit{All} \\
        \midrule
        BEVDepth$^\dagger$ & 38.1 & 40.9 & 24.1 & 15.0 & 26.7 \\
        \rowcolor{gray9} \textbf{X$^{3}$KD}$_{\mathrm{modal}}$ & 39.1 & 41.9 & 24.6 & 17.3 & 27.9 \\
        \rowcolor{gray8} \textbf{X$^{3}$KD}$_{\mathrm{all}}$ & \textbf{39.6} & \textbf{43.4} & \textbf{25.4} & \textbf{17.7} & \textbf{28.8} \\
        \bottomrule
    \end{tabular}
    \vspace{-6pt}
    \caption{\textbf{Performance comparison on the Waymo dataset.} We compare X$^{3}$KD to our re-implemented baseline BEVDepth$^{\dagger}$~\cite{li2022bevdepth}.}
    \label{tab:sota-waymo}
\end{table}

\subsection{Experimental Setup}
\vspace{-3pt}
\label{sec:exp-setup}

X$^3$KD is implemented using mmdetection3d~\cite{mmdet3d2020} and PyTorch~\cite{paszke2019pytorch} libraries and trained on 4 NVIDIA A100 GPUs.\footnote{We use mmdetection3d v1.0, Python 3.8, PyTorch 1.11, CUDA 11.3} Here, we describe our main setup on nuScenes while more details are provided in the supplementary.
\par 
\textbf{Datasets}:
Similar to most recent works~\cite{bai2022transfusion, li2022bevdepth, li2022bevformer, li2022bevstereo, yin2021center}, we evaluate on the nuScenes and Waymo benchmark datasets. The nuScenes dataset~\cite{caesar2020nuscenes} contains 28K, 6K, and 6K samples for training, validation, and test, respectively. We use data from a LiDAR sensor and 6 cameras with bounding box annotations for 10 classes. For the Waymo dataset~\cite{sun2020waymo}, we use the data from a LiDAR sensor and 5 cameras with annotations for cars, pedestrians, and cyclists. It provides 230K annotated frames from 798, 202, and 150 sequences for training, validation, and test, respectively. 
\par 
\textbf{Evaluation Metrics}:
For nuScenes, we employ the officially defined $\textit{mAP}$ and $\textit{NDS}$ metrics. The $\textit{NDS}$ metric considers $\textit{mAP}$ as well as true positive ($\textit{TP}$) metrics $\mathbb{TP} = \left\lbrace\textit{mATE}, \textit{mASE}, \textit{mAOE}, \textit{mAVE}, \textit{mAAE}\right\rbrace$ for translation, scale, orientation, velocity, and attribute, respectively, \ie, $\textit{NDS} = \frac{1}{10}\left(5\cdot \textit{mAP}\right) + \sum_{\textit{TP}\in\mathbb{TP}} 1 - \min\left(1, \textit{TP}\right)$. 
For Waymo, we employ the official metrics of the camera-only 3D object detection track~\cite{hung2022let}: The $\textit{LET-3D-AP}$ calculates average precision after longitudinal error correction, while $\textit{LET-3D-APL}$ also penalizes the longitudinal error.
\par 
\begin{table*}[t]
\captionsetup{font=small, belowskip=-14pt}
	\footnotesize
    \centering
    \setlength{\tabcolsep}{8.7pt}
    \def\arraystretch{0.97}%
    \begin{tabular}{l|cccc|ccccc|cc}
        \toprule
        \textit{Model} & 
        X-OD & X-FD & X-AT & X-IS & 
        \cellcolor[HTML]{96bbce}\textit{mATE}$\downarrow$ & 
    	\cellcolor[HTML]{fb9a99}\textit{mASE}$\downarrow$ & 
     	\cellcolor[HTML]{fc4538}\textit{mAOE}$\downarrow$ & 
    	\cellcolor[HTML]{fdbf6f}\textit{mAVE}$\downarrow$ & 
    	\cellcolor[HTML]{ff7f00}\textit{mAAE}$\downarrow$ & 
    	\cellcolor[HTML]{ab9ac0}\textit{mAP}$\uparrow$ & 
    	\cellcolor[HTML]{a5eb8d}\textit{NDS}$\uparrow$ \\
    	\midrule
        BEVDepth$^\dagger$ & \xm & \xm & \xm & \xm & 0.636 & 0.272 & 0.493 & 0.499 & 0.198 & 35.9 & 47.2 \\
        X-OD & \ch & \xm & \xm & \xm & 0.642 & 0.278 & \textbf{0.456} & \textbf{0.338} & \textbf{0.188} & 35.7 & 48.7 \\
        X-FD & \xm & \ch & \xm & \xm & 0.644 & 0.276 & 0.479 & 0.361 & 0.200 & 36.1 & 48.5 \\
        X-AT & \xm & \xm & \ch & \xm & 0.648 & 0.277 & 0.492 & 0.354 & \underline{0.192} & 35.5 & 48.1 \\
        \rowcolor{gray9} \textbf{X$^{3}$KD}$_{\mathrm{modal}}$ & \ch & \ch & \ch & \xm & \underline{0.632} & \underline{0.271} & \textbf{0.456} & \underline{0.342} & 0.203 & 36.8 & 49.4 \\
        \rowcolor{gray85} X-IS & \xm & \xm & \xm & \ch & 0.635 & 0.273 & \underline{0.462} & 0.350 & 0.204 & \underline{38.7} & \underline{50.1} \\
        \rowcolor{gray8}  \textbf{X$^{3}$KD}$_{\mathrm{all}}$ & \ch & \ch & \ch & \ch & \textbf{0.615} & \textbf{0.269} & 0.471 & 0.345 & 0.203 & \textbf{39.0} & \textbf{50.5} \\
        \midrule
        LiDAR Teacher & \nap & \nap & \nap & \nap & 0.301 & 0.257 & 0.298 & 0.256 & 0.195 & 59.0 & 66.4 \\
        \bottomrule
    \end{tabular}
    \vspace{-6pt}
    \caption{\textbf{Ablation study of X$^{3}$KD on the nuScenes validation set}: We incrementally add our proposed cross-modal feature distillation (X-FD), adversarial training (X-AT) and output distillation (X-OD) as well as our cross-task instance segmentation distillation (X-IS). All X$^{3}$KD variants in the top part are solely based on multi-camera images during inference. Best numbers in boldface, second best underlined.}
    \label{tab:ablation-main}
\end{table*}

\textbf{Network Architecture and Training}:
For a fair comparison, our network architecture follows previous works~\cite{huang2022bevdet4d, jiang2022polarformer, li2022bevdepth, li2022bevformer, li2022bevstereo, wang2022sts}. We consider the ResNet-50-based setting with a resolution of $256\times 704$ and the ResNet-101-based setting with resolutions of $512\times 1408$ or $640\times 1600$. Further network design choices are adopted from~\cite{huang2022bevdet4d}. We train all models for $24$ epochs using the CBGS training strategy~\cite{zhu2019class}, a batch size of $16$ and AdamW~\cite{loshchilov2018decoupled} with an initial learning rate of $2\cdot 10^{-4}$. The loss weights are set to $\lambda^{\text{GT}}=1$, $\lambda^{\text{X-FD}}=10$, $\lambda^{\text{X-AT}}=10$, $\lambda^{\text{X-OD}}=1$, and $\lambda^{\text{X-IS}}=1$ while the thresholds are set to $\alpha^{\text{3D-bbox}}=0.6$ and $\alpha^{\text{2D-bbox}}=0.2$. Our LiDAR teacher is based on the CenterPoint architecture \cite{yin2021center} and the TransFusion training schedule~\cite{bai2022transfusion}. The supplementary contains further explanations, hyperparameter studies, and configurations for the Waymo dataset.
\vspace{-2mm}
\subsection{State-of-the-art Comparisons}
\vspace{-3pt}
\label{sec:sota-comp}
We perform a comparison of X$^{3}$KD with all contributions, \ie, X$^{3}$KD$_{\text{all}}$, to other SOTA methods in Table~\ref{tab:sota-nuscenes}. In the ResNet-50-based setting, our model achieves the best results with scores of $39.0$ and $50.5$ in \textit{mAP} and \textit{NDS}, respectively. In the high-resolution ResNet-101-based setting, our model achieves SOTA scores of $46.1$ and $56.7$. \textit{At this resolution, we outperform all previous SOTA methods in all considered metrics and outperform the second best result by 2.9 points in \textit{mAP} and 2.5 points in \textit{NDS}}. To explicitly show that our method improves on top of current SOTA baselines, we retrain our strongest baseline among published works, \ie, BEVDepth~\cite{li2022bevdepth}, in our code framework, dubbed BEVDepth$^\dagger$. At all resolutions, we are able to closely reproduce the reported results and improve by about 3 points in both \textit{mAP} and \textit{NDS} upon them. \textit{On the test set, we outperform the second best approach PolarFormer~\cite{jiang2022polarformer} by $1.8$ points in terms of the main NDS metric} and achieve best results in 5 out of 7 metrics. We also show results for BEVDepth$^\dagger$ and X$^3$KD variants on the Waymo dataset in Table~\ref{tab:sota-waymo}. As on nuScenes, our X$^{3}$KD$_{\text{all}}$ model clearly outperforms the baseline in all metrics.
\begin{table}[t]
	\captionsetup{font=small, belowskip=-12pt}
    \centering
    \footnotesize
    \def\arraystretch{0.97}%
    \setlength{\tabcolsep}{2.7pt}
    \begin{tabular}{l|ccc|cccc}
        \toprule
        \textit{Model} &\textit{Dist.} & \textit{Weight} & \textit{w/o GT} & 
        \cellcolor[HTML]{fc4538}\textit{mAOE}$\downarrow$ & 
        \cellcolor[HTML]{fdbf6f}\textit{mAVE}$\downarrow$ & 
        \cellcolor[HTML]{ab9ac0}\textit{mAP}$\uparrow$ & 
        \cellcolor[HTML]{a5eb8d}\textit{NDS}$\uparrow$ \\
        \midrule
        BEVDepth$^\dagger$ & \xm & \xm & \xm & 0.493 & 0.499 & \textbf{35.9} & 47.2 \\
        & \ch & \xm & \xm & 0.477 & 0.342 & 35.6 & 48.5 \\
        \rowcolor{gray85} X-OD & \ch & \ch & \xm & \textbf{0.456} & \textbf{0.338} & 35.7 & \textbf{48.7} \\
        \midrule
        & \ch & \xm & \ch & 1.090 & 0.972 & 36.1 & 35.3 \\
        \rowcolor{gray85} X-OD$_{\text{w/o GT}}$ & \ch & \ch & \ch & \textbf{0.724} & \textbf{0.570} & \textbf{36.5} & \textbf{43.7} \\
        \bottomrule
    \end{tabular}
    \vspace{-6pt}
    \caption{\textbf{Ablation study on cross-modal output distillation (X-OD) on the nuScenes validation set.} We show the effect of weighing the regression loss in (\ref{eq:x-od}) by the teacher output probabilities $\langle\tilde{\bm{b}}_s^{\text{cls}} \rangle$ (Weight) during distillation (Dist.). We also show that our method can be trained without annotations (w/o GT).}
    \label{tab:ablation-output-distill}
\end{table}

\subsection{Method Ablation Studies}
\vspace{-3pt}
\label{sec:ablation}

\textbf{Effectiveness of the Proposed Components}:
We incrementally add our contributions in Table~\ref{tab:ablation-main} and evaluate them in terms of \textit{NDS} and \textit{mAP}. First, we individually add X-OD, X-FD, and X-AT. For all three components, there is an improvement in the \textit{NDS} metric, while the \textit{mAP} metric remains similar or slightly worse. Adding all three components (X$^{3}$KD$_{\mathrm{modal}}$) gives a clear improvement over the baseline as well as applying each component individually. Particularly, we observe that the \textit{additional cross-modal supervision improves bounding box velocity estimation from multi-camera input} as can be seen by the apparent improvement in the \textit{mAVE} metric. Using X-IS, surprisingly gives an even more substantial improvement. This might indicate that \textit{supervision in BEV cannot completely compensate for the lack of rich features in PV}. Finally, adding all components together to \textit{our proposed} X$^{3}$KD$_{\mathrm{all}}$ \textit{model clearly outperforms all other variants} in terms of the main \textit{NDS} and \textit{mAP} metrics and is best in 4 out of 7 metrics in Table~\ref{tab:ablation-main}.
\par
\begin{table}[t]
	\captionsetup{font=small, belowskip=-14pt}
    \centering
    \footnotesize
    \def\arraystretch{0.97}%
    \setlength{\tabcolsep}{3pt}
    \begin{tabular}{l|c|c|cc|cc}
        \toprule
        \textit{Model} & \textit{Student} & \textit{Teacher} & \textit{Pre.} & \textit{Dist.} & \cellcolor[HTML]{ab9ac0}\textit{mAP}$\uparrow$ & 
        \cellcolor[HTML]{a5eb8d}\textit{NDS}$\uparrow$ \\
        & \textit{Backbone} & \textit{Backbone} & & & & \\
        \midrule
        \multirow{1}{*}{BEVDepth$^\dagger$} & \multirow{1}{*}{ResNet-50} & \nap & \xm & \xm & 35.9 & 47.2 \\
        \midrule
        & \multirow{1}{*}{ResNet-50} & ResNet-50 & \xm & \ch & 36.4 & 48.8 \\
        & ResNet-50 & \nap & \ch & \xm & 37.7 & 49.5 \\ 
        \rowcolor{gray85} X-IS & ResNet-50 & ResNet-50 & \ch & \ch & \textbf{38.7} & \textbf{50.1} \\
        \rowcolor{gray85} X-IS & ResNet-50 & ConvNeXt-T & \ch & \ch & 38.5 & 49.9 \\
        \midrule
        & \multirow{1}{*}{ConvNeXt-T} & \nap & \xm & \xm & 38.3 & 50.8 \\
        & ConvNeXt-T & ResNet-50 & \xm & \ch & \textbf{38.9} & \textbf{51.4} \\
        \bottomrule
    \end{tabular}
    \vspace{-6pt}
    \caption{\textbf{Ablation study on cross-task instance segmentation distillation (X-IS) on the nuScenes validation set.} We evaluate the effect of using pre-trained weights (Pre.) and knowledge distillation (Dist.) as well as different teacher/student backbones.}
    \label{tab:ablation-inst-seg}
\end{table}
\begin{figure*}
    \captionsetup{font=small, belowskip=-12pt}
    \centering
    \includegraphics[width=1.0\linewidth]{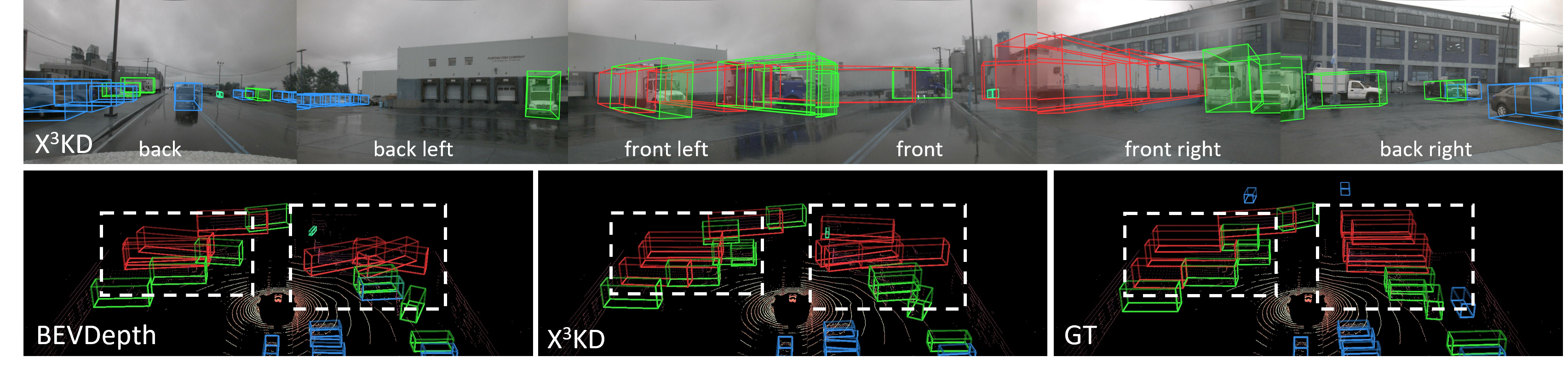}
    \vspace{-25pt}
    \caption{\textbf{Qualitative results on nuScenes}: We show the multi-camera input (top) and bounding box visualizations (bottom). We compare ResNet-101-based X$^3$KD$_\text{all}$ to BEVDepth$^\dagger$ and the ground truth (GT) for a resolution of $640 \times 1600$. Best viewed on screen and in color.}
    \label{fig:qualitative}
\end{figure*}
\textbf{Cross-Modal Output Distillation (X-OD)}:
We provide insights into our X-OD design in Table~\ref{tab:ablation-output-distill}. In the top part, we observe that models trained with output distillation improve over the baseline in terms of \textit{NDS} and that the confidence-based weighting is particularly effective for orientation (\textit{mAOE}) and velocity (\textit{mAVE}) prediction. Further, we train the multi-camera 3DOD without using annotations (Table~\ref{tab:ablation-output-distill}, bottom part) solely from KD. In this setting, the weighting yields even more significant improvements in particular in terms of the \textit{NDS} metric. Also, \textit{the} X-OD$_{\text{w/o GT}}$ \textit{model surprisingly outperforms the model variants trained with annotations in terms of the \textit{mAP} metric.} This promising result indicates that future work might be able to use large-scale pre-training with KD on unlabelled data for further performance improvements. 
\par
\textbf{Cross-task Instance Segmentation Distillation (X-IS)}:
Ablations on our X-IS design are shown in Table~\ref{tab:ablation-inst-seg}. We observe that initialization of the backbone with weights from a pre-trained instance segmentation as well as cross-task distillation, improves the baseline's result. Combining both aspects to X-IS yields the best result in both \textit{mAP} and \textit{NDS}. Using a different teacher model based on ConvNeXt-T yields similarly good results and shows that the feature extraction architectures of the instance segmentation teacher and the multi-camera 3DOD student do not need to match. Also, knowledge can be distilled from a simple ResNet-50-based model into a more sophisticated architecture such as ConvNeXt-T (bottom part of Table~\ref{tab:ablation-inst-seg}). Overall, \textit{cross-task distillation can improve performance without requiring an additional pre-training step}.
\subsection{Method Analysis}
\begin{figure}
    \captionsetup{font=small, belowskip=-14pt}
    \centering
    \vspace{-5pt}
    \includegraphics[width=0.95\linewidth]{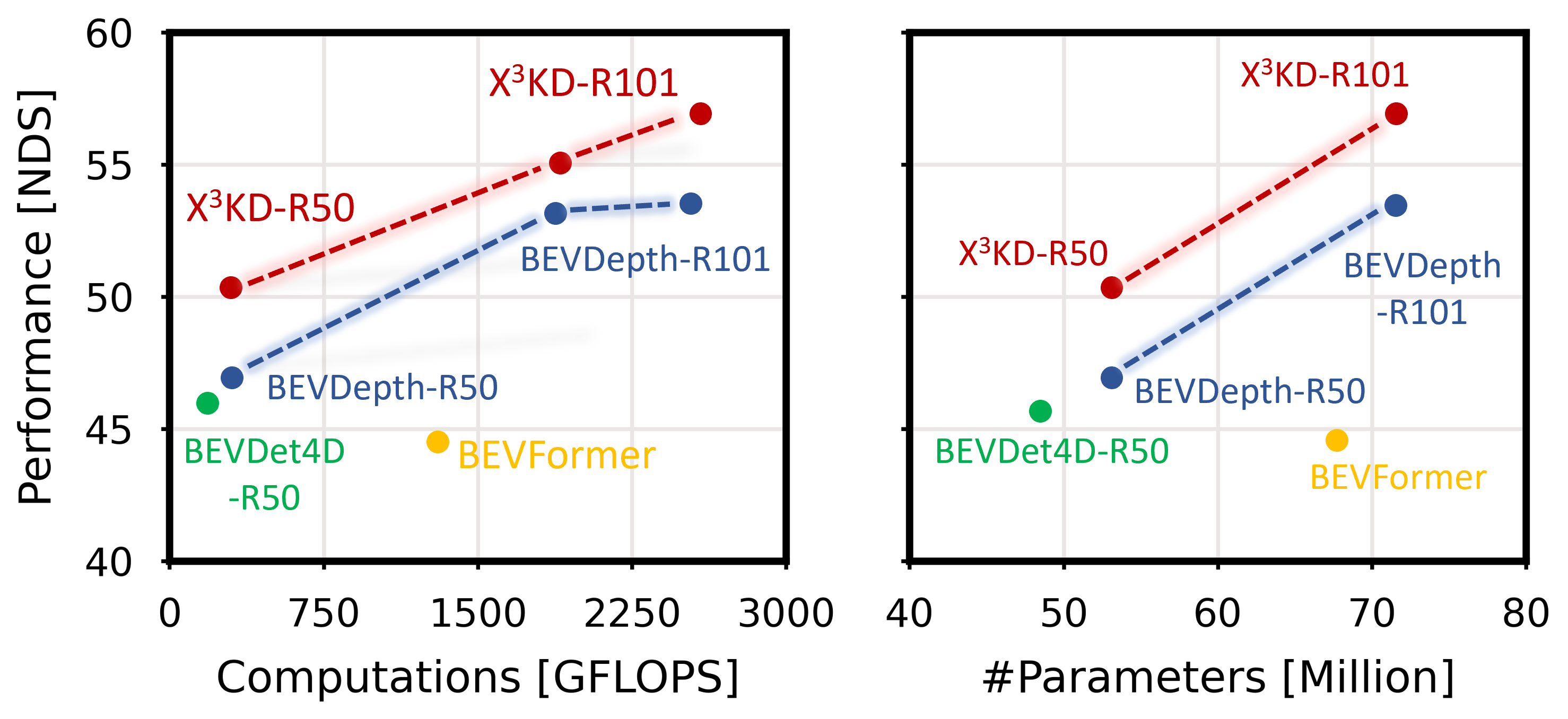}
    \vspace{-6pt}
    \caption{\textbf{Complexity Analysis} of X$^3$KD in comparison to BEVDepth~\cite{li2022bevdepth}, BEVDet4D~\cite{huang2022bevdet4d}, and BEVFormer~\cite{li2022bevformer}.}
    \label{fig:complexity}
\end{figure}

\vspace{-3pt}
\label{sec:analysis}
\begin{table}[t]
	\captionsetup{font=small, belowskip=-8pt}
    \centering
    \footnotesize
    \def\arraystretch{0.9}%
    \setlength{\tabcolsep}{5.0pt}
    \begin{tabular}{l|cc|cc|cc}
        \toprule
        \textit{Model} & \textit{RADAR} & \textit{Cam.} & \multicolumn{2}{c|}{Validation} & \multicolumn{2}{c}{Test} \\
        & \textit{Input} & \textit{Input} & \cellcolor[HTML]{ab9ac0}\textit{mAP}$\uparrow$ & \cellcolor[HTML]{a5eb8d}\textit{NDS}$\uparrow$ & \cellcolor[HTML]{ab9ac0}\textit{mAP}$\uparrow$ & \cellcolor[HTML]{a5eb8d}\textit{NDS}$\uparrow$ \\
        \midrule
		RADAR only & \ch & \xm & 12.9 & 13.0 & - & - \\
		\rowcolor{gray85} \textbf{X$^{3}$KD}$_{\mathrm{modal}}$ & \ch & \xm & \textbf{17.7} & \textbf{23.5} & - & - \\
		\midrule
		Fusion only & \ch & \ch & 38.9 & 51.0 & 40.2 & 52.3 \\
		\rowcolor{gray85} \textbf{X$^{3}$KD}$_{\mathrm{modal}}$ & \ch & \ch & \textbf{42.3} & \textbf{53.8} & \textbf{44.1} & \textbf{55.3} \\
        \bottomrule
    \end{tabular}
    \vspace{-3pt}
    \caption{\textbf{Generalization of our method to RADAR:} We distill knowledge from a LiDAR-based 3DOD into a RADAR-based and a RADAR-camera fusion-based 3DOD model. For RADAR-based models, we report the mAP just for the car class as these models underperform on other classes due to the point cloud sparsity.}
    \label{tab:generalization-radar}
\end{table}
\textbf{Performance-Complexity Trade-off}:
We analyze our method's efficiency compared to state-of-the-art methods~\cite{li2022bevformer,li2022bevdepth,huang2022bevdet4d} in Fig.~\ref{fig:complexity}. We compare to reimplementations of BEVDepth~\cite{li2022bevdepth} and BEVDet4D~\cite{huang2022bevdet4d} as well as reported results of BEVFormer~\cite{li2022bevformer}. All reported models are ResNet-50-based or ResNet-101-based to ensure that a better trade-off cannot be attributed to a more efficient backbone. We observe that X$^3$KD (red curve) outperforms BEVDepth (blue curve) at equal complexity due to the improved supervision from KD. Also, compared to BEVDet4D and BEVFormer a better trade-off can be observed, likely because of the absence of LiDAR supervision in BEVDet4D and the complex Transformer model in BEVFormer. Accordingly, \textit{our results show that X$^3$KD achieves a better complexity-performance trade-off than current state-of-the-art methods.}
\par
\textbf{Qualitative Results}:
We further show qualitative results of X$^3$KD and BEVDepth in Fig.~\ref{fig:qualitative}. As highlighted by the white boxes, X$^3$KD detects and places objects more accurately in the scene. In particular, the recognition of objects and the prediction of their orientation shows improved characteristics in the X$^3$KD output, which is coherent with a better quantitative performance of X$^3$KD in Table~\ref{tab:ablation-main}. Further qualitative results are given in the supplementary.
\subsection{Generalization to RADAR}
\vspace{-3pt}
\label{sec:gen-to-radar}
We also generalize X$^3$KD to RADAR-based and camera-RADAR fusion-based models. For RADAR-based models, we cannot apply cross-task KD from the instance segmentation teacher. Hence, we only use the cross-modal KD contributions, \ie, X$^{3}$KD$_{\mathrm{modal}}$. Our results on the nuScenes validation set show that X$^{3}$KD$_{\mathrm{modal}}$ significantly enhances the performance in both settings. Notably, the transfer from camera to RADAR was straightforward as we achieved the reported improvements without requiring tuning of hyperparameters. Further, we evaluate our fusion-based X$^{3}$KD$_{\mathrm{modal}}$ model on the nuScenes test set, where \textit{we outperform all other Camera-RADAR, fusion-based models, hence setting the state-of-the-art result}.

\section{Conclusions}
\vspace{-4pt}
We proposed X$^3$KD, a KD framework for multi-camera 3DOD. By distilling across tasks from an instance segmentation teacher and across modalities from a LiDAR-based 3DOD teacher into different stages of a multi-camera 3DOD student, we show that the model performance can be enhanced without inducing additional complexity during inference. We evaluated X$^3$KD on the nuScenes and Waymo datasets, outperforming previous approaches by 2.9\% \textit{mAP} and 2.5\% \textit{NDS}.
The transferability to other sensors, such as RADAR, and the possibility to train 3DOD models without annotations further demonstrate X$^3$KD's effectiveness. Combining these two findings could be used in future applications to train 3DOD models for arbitrary sensors, requiring only a LiDAR-based 3DOD model.

{\small
\bibliographystyle{ieee_fullname}
\bibliography{bib/bibliography}}
\end{document}